\documentclass[letterpaper]{article} 
\usepackage{aaai2026}  
\usepackage{times}  
\usepackage{helvet}  
\usepackage{courier}  
\usepackage[hyphens]{url}  
\usepackage{graphicx} 
\usepackage{natbib}  
\usepackage{caption} 
\usepackage{algorithm}
\usepackage{algorithmic}

\usepackage{newfloat}
\usepackage{listings}

\usepackage{amsmath,amsfonts}
\usepackage{xcolor}
\usepackage{multirow}
\usepackage{array}
\usepackage{booktabs}
\usepackage{hhline}
\usepackage[table]{xcolor}
\usepackage{cite}
\usepackage{pifont} 
\usepackage{tabularx}

\DeclareCaptionStyle{ruled}{labelfont=normalfont,labelsep=colon,strut=off} 
\floatstyle{ruled}
\newfloat{listing}{tb}{lst}{}
\floatname{listing}{Listing}
\title{SSR: Semantic and Spatial Rectification for CLIP-based Weakly Supervised Segmentation}
\author{
    Xiuli Bi\textsuperscript{\rm 1},
    Die Xiao\textsuperscript{\rm 1},
    Junchao Fan\textsuperscript{\rm 1}\thanks{Corresponding author},
    Bin Xiao\textsuperscript{\rm 1, \rm2},
}
\affiliations{
    \textsuperscript{\rm 1}Chongqing Key Laboratory of Image Cognition,\\
Chongqing University of Posts and Telecommunications, Chongqing, China\\
    \textsuperscript{\rm 2}Jinan Inspur Data Technology Co., Ltd., Jinan, China
    \\
    bixl@cqupt.edu.cn, s230201127@stu.cqupt.edu.cn, fanjc@cqupt.edu.cn, xiaobin@cqupt.edu.cn
}

\usepackage{bibentry}

\begin{document}

\maketitle

\begin{abstract}
In recent years, Contrastive Language-Image Pretraining (CLIP) has been widely applied to Weakly Supervised Semantic Segmentation (WSSS) tasks due to its powerful cross-modal semantic understanding capabilities. This paper proposes a novel Semantic and Spatial Rectification (SSR) method to address the limitations of existing CLIP-based weakly supervised semantic segmentation approaches: over-activation in non-target foreground regions and background areas. Specifically, at the semantic level, the Cross-Modal Prototype Alignment (CMPA) establishes a contrastive learning mechanism to enforce feature space alignment across modalities, reducing inter-class overlap while enhancing semantic correlations, to rectify over-activation in non-target foreground regions effectively; at the spatial level, the Superpixel-Guided Correction (SGC) leverages superpixel-based spatial priors to precisely filter out interference from non-target regions during affinity propagation, significantly rectifying background over-activation. Extensive experiments on the PASCAL VOC and MS COCO datasets demonstrate that our method outperforms all single-stage approaches, as well as more complex multi-stage approaches, achieving mIoU scores of 79.5\% and 50.6\%, respectively.
\end{abstract}


\section{Introduction}

The goal of WSSS is to generate high-quality pseudo labels by using annotations such as points~\cite{bearman2016s}, scribbles\cite{vernaza2017learning}, bounding boxes~\cite{lee2021bbam}, or image-level labels\cite{pinheiro2015image, lee2022weakly}, thereby addressing the issues of high creation cost and time consumption associated with pixel-level annotations in fully supervised semantic segmentation~\cite{long2015fully}. Among these annotation, image-level labels are the most popular yet most challenging form of annotation. In this work, we study WSSS with image-level labels.

Current WSSS methods typically follow a three-stage pipeline: 1) training a classification network to generate initial Class Activation Maps (CAM)~\cite{zhou2016learning}, 2) refining these CAM~\cite{ahn2019weakly}, and 3) producing pseudo-labels for segmentation model~\cite{xie2021segformer} training. While conventional CNN-based CAM methods~\cite{guo2022cmt,he2024progressive}suffer from under-activation (focusing only on the most discriminative object parts) due to limited receptive fields, recent Vision Transformer (ViT)-based approaches~\cite{qin2022activation} demonstrate superior performance by capturing global contexts through Multi-Head Self-Attention (MHSA). However, ViT architectures introduce a new challenge of over-activation (erroneously activating background regions), which current research~\cite{yoon2024class, yoon2024diffusion} addresses through improved positional encoding~\cite{wu2021rethinking} or CNN-ViT hybrid architectures.

\begin{figure}[t]
\centering
\includegraphics[width=0.98\columnwidth]{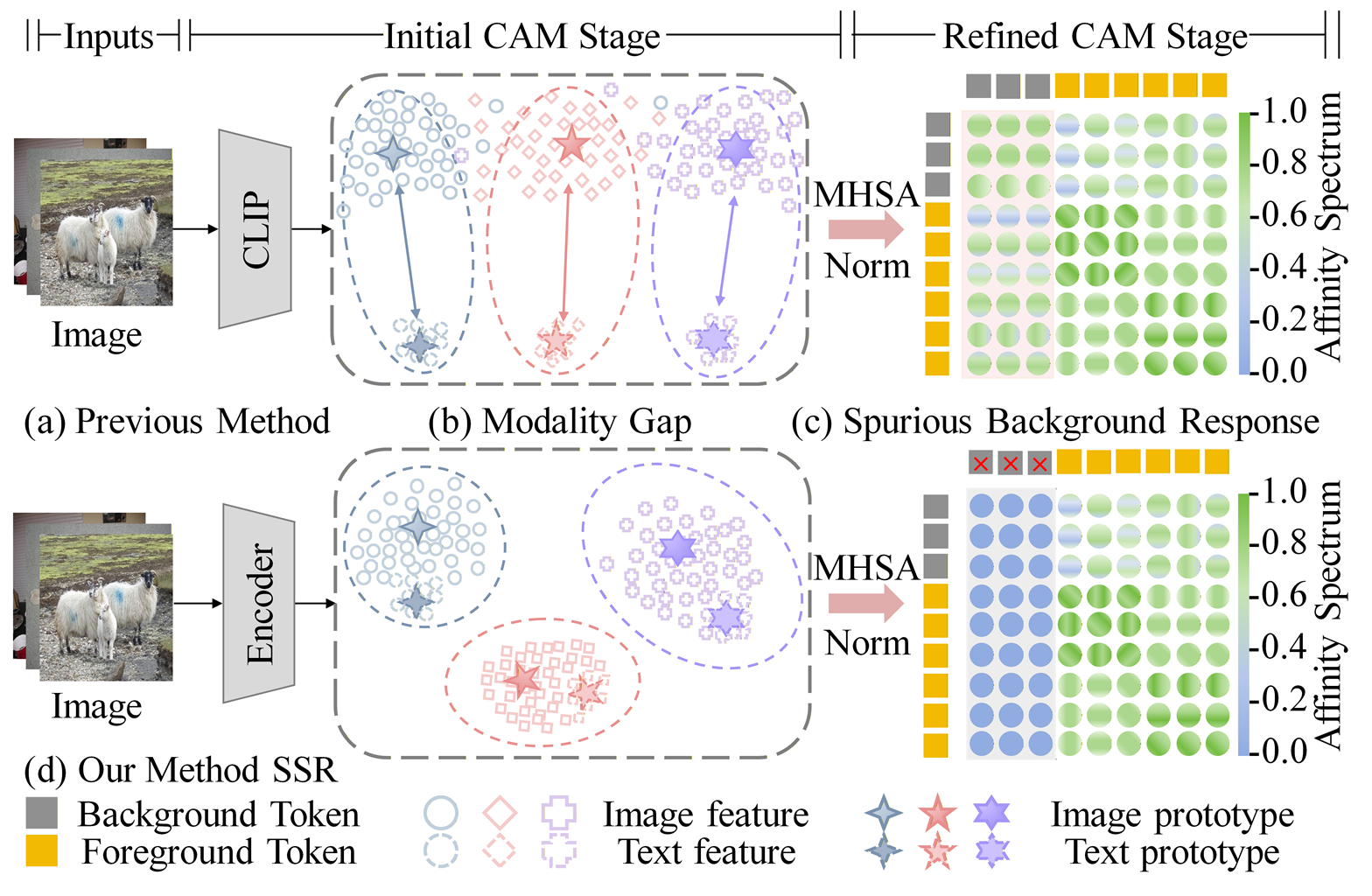}
\caption{Our motivation. (a) Previous methods exhibit inherent limitations. (b) The initial CAM stage suffers from a modality gap, where the visual feature space faces dual challenges of intra-class dispersion and inter-class overlap. (c) The refined CAM stage is plagued by spurious background responses, as affinity estimation is corrupted by background noise. (d) To address these issues, we propose SSR.}
\label{fig1}
\end{figure}

Recently, researchers have begun to explore the integration of CLIP~\cite{radford2021learning} into WSSS~\cite{yang2024foundation, yang2025exploring}. They employ gradient-based methods~\cite{selvaraju2017grad} to generate initial CAM, which exhibit significantly superior performance compared to traditional CNN and ViT solutions, leading to their widespread adoption in WSSS applications.

As a multimodal model, CLIP projects images and text into a shared representation space where matched image-text embeddings are pulled closer while mismatched ones are pushed apart. However, in downstream applications, CLIP still faces the issue of over activation in non-target foreground regions due to insufficient semantic alignment between image and text modalities, which stems from its inherent modality gap~\cite{liang2022mind}—the fundamental differences in the underlying feature distributions of the two modalities, as shown in Figure \ref{fig1}. Specifically, visual features primarily focus on low-level patterns, such as colors and shapes, whereas textual features emphasize higher-level, abstract semantics. To address this semantic misalignment issue, existing improvement methods primarily focus on optimizing the quality of text prompts. However, these methods~\cite{yang2024foundation, yang2025exploring} merely modify the text-side representations without fundamentally bridging the cross-modal representational gap between visual and textual modalities. Consequently, they fail to resolve the activation errors caused by semantic misalignment completely.

During the feature refinement process, the abnormally high affinity values between background regions and target regions often lead to erroneous activation of background areas, consequently triggering the spurious background response issue (as illustrated in Figure \ref{fig1}). To address this problem, existing research primarily follows two solution approaches. The first employs a multi-stage iterative optimization strategy~\cite{zhang2021affinity} to eliminate background noise through progressively alternating training mechanisms. The second implements affinity matrix constraints~\cite{ru2022learning} by applying threshold truncation or introducing constrained loss functions to enhance the reliability of supervision signals. However, these methods still suffer from interference by low-level features and the problem of global context confusion.

In this work, we propose a Semantic and Spatial Rectification to tackle the semantic misalignment caused by modality gaps and the affinity noise from spurious background response. At the semantic level, our approach aims to: 1) preserve cross-modal semantic consistency in features, and 2) ensure pixels of the same category exhibit similar representations in feature space. To achieve this, we design Cross-Modal Prototype Alignment, which enhances inter-modal alignment and feature discriminability, providing a more reliable basis for pseudo-label generation. To further enhance the quality of initial CAM for more precise pseudo-labels at the spatial level, we propose Superpixel Guided Correction. SGC addresses potential erroneous associations between target and background patch tokens by incorporating superpixel guidance, which effectively strengthens foreground semantic consistency while suppressing interference from background noise.

The main contributions of our work are listed as follows:
\begin{itemize}
\item We proposes a Semantic and Spatial Rectification method to address the over-activation issues concerning non-target foreground regions and background areas in CLIP-based weakly supervised semantic segmentation.
\item The Cross-Modal Prototype Alignment establishes a contrastive learning mechanism at the semantic level to align the feature representations across different modalities in the feature space.
\item Superpixel Guided Correction operates at the spatial level to rectify ViT's affinity matrix by leveraging superpixel segmentation, enabling local affinity optimization.
\item Extensive experiments on PASCAL VOC and MS COCO demonstrate that our method significantly outperforms recent state-of-the-art approaches.
\end{itemize}

\section{Related Work}
\subsection{Weakly Supervised Semantic Segmentation}
To address the limitation that CAM highlight only the most discriminative object regions, researchers have proposed various innovative solutions. Erasing~\cite{kweon2021unlocking}, cross-image mining~\cite{sun2020mining, li2021group}, self-supervised learning~\cite{wang2020self, chen2022self}, and adversarial attack~\cite{lee2021anti} approaches have enhanced CAM coverage from different perspectives. Some works adopt prototype learning frameworks: PSDPM~\cite{zhao2024psdpm} activates more secondary discriminative pixels using class prototypes, while FPR~\cite{chen2023fpr} constructs category-specific positive and negative prototypes.The emergence of ViTs has demonstrated promising localization capabilities. A2GNN~\cite{zhang2021affinity} builds semantic graph structures through affinity convolution, while AFA~\cite{ru2022learning} leverages the MHSA mechanism to explore pixel-level semantic relationships. Other approaches focus on position encoding: CTI~\cite{yoon2024class} injects category-specific tokens from images, and MCTformer~\cite{xu2022multi} optimizes class-specific attention maps through inter-block pairwise affinity computation.

\begin{figure*}[t] 
\centering
\includegraphics[width=0.99\textwidth]{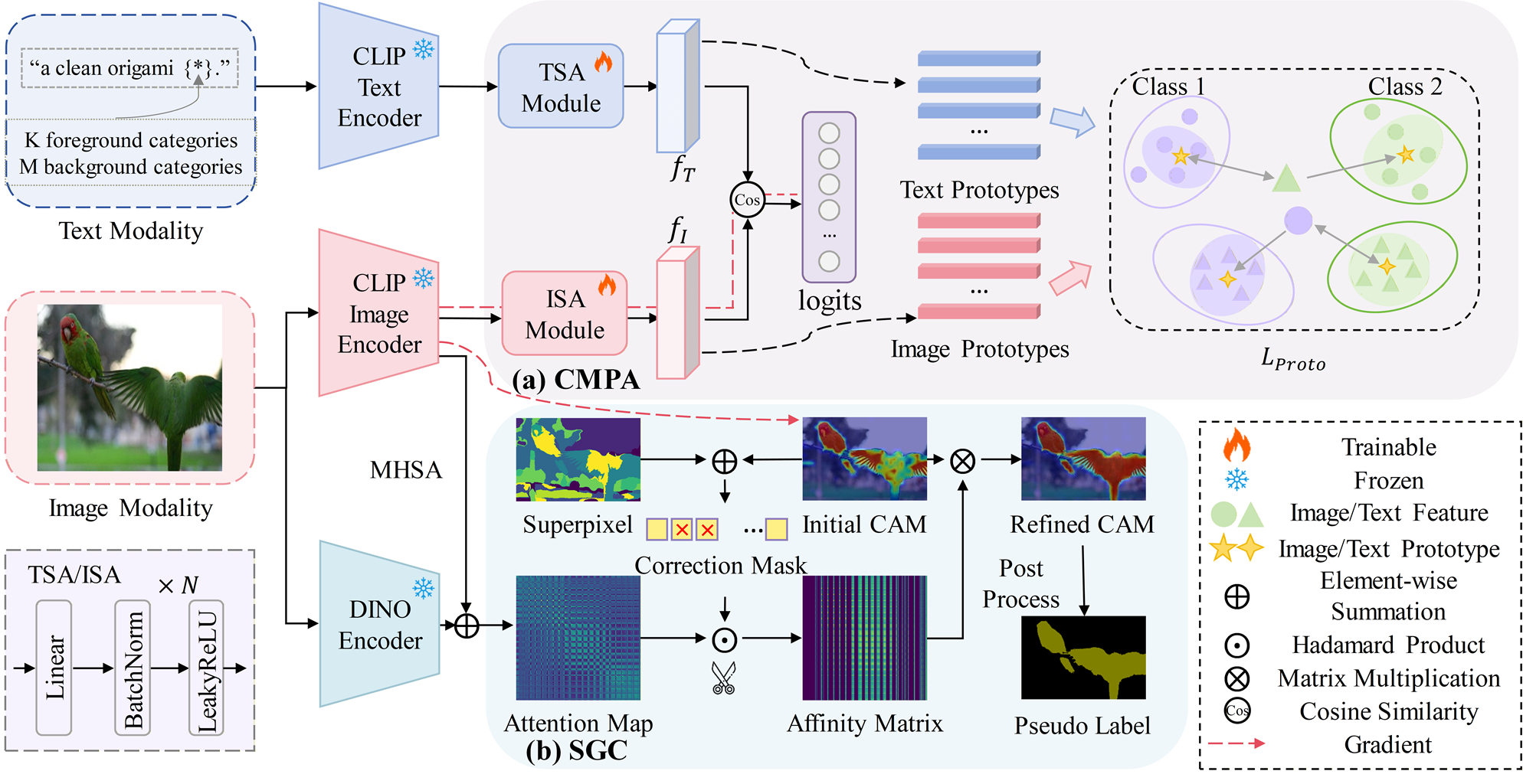} 
\caption{Overview of our SSR.We propose two novel components to address the key challenges of modality gap and erroneous activation: the CMPA and SGC. (a) The CMPA utilizes cross-modal prototype contrastive learning to establish precise matching relationships between visual features and textual prototypes in a shared embedding space, thereby effectively alleviating class confusion. (b) The SGC utilizes local spatial consistency priors derived from superpixel clustering to selectively filter the feature affinity matrix selectively, eliminating erroneous cross-region propagation and guiding the feature refinement process toward semantically consistent directions, thereby significantly suppressing background over-activation phenomena.}
\label{fig2}
\end{figure*}

\subsection{Contrastive Language-Image Pre-training}
In the field of WSSS, CLIP has attracted significant attention due to its exceptional performance. CLIMS~\cite{xie2022clims} employs contrastive learning in both foreground and background regions to perform cross-lingual image matching. CLIP-ES~\cite{lin2023clip} employs carefully designed manual text prompts and utilizes the softmax function to generate GradCAM. WeakCLIP~\cite{zhu2025weakclip} transforms the WSSS task into a continuous text-to-image matching problem, effectively leveraging vision-language pre-training knowledge. WeCLIP~\cite{zhang2024frozen} directly adopts the CLIP visual encoder to accomplish segmentation tasks. However, these methods all face the inherent modality gap~\cite{liang2022mind} issue in multimodal models, leading to inconsistent semantic alignment between visual and textual features. To address this challenge, researchers have proposed various innovative solutions. The FMA~\cite{yang2024foundation} method optimizes textual feature representation by designing learnable prompts separately for classification and segmentation tasks. ExCEL~\cite{yang2025exploring} enriches text prompt information by generating fine-grained category descriptions with the help of large language models. VPL~\cite{xu2025toward} employs a gradient descent approach to learn category-specific visual prototypes in the visual space, replacing traditional text prototypes to more accurately capture the features of semantic target regions.

\section{Methodology}
\subsection{Framework Overview}
Due to the significant discrepancy between textual and visual modalities, conventional text-feature-optimized alignment methods often struggle to establish precise pixel-level semantic correspondences, leading to erroneous activation in non-target foreground regions. During the refinement of CAM, background over-activation frequently occurs due to noise in affinity matrix. 
To systematically address these challenges, the proposed SSR Framework innovatively conducts collaborative modeling and joint optimization from dual dimensions of semantic understanding and spatial relationships, thereby achieving more accurate pixel-level semantic alignment across modalities.

Figure \ref{fig2} illustrates the overall framework of SSR. The cross-modal input consists of an image modality $I$ and a text modality $T$, where $T$ comprises $K$ foreground categories (class labels $Y$={1,2,...,C}) and $M$ background categories derived from CLIP-ES\cite{lin2023clip}. At the semantic level, Cross-Modal Prototype Alignment significantly reduces the modality gap through contrastive learning between image and text prototypes, achieving compact cross-modal feature alignment. At the spatial level, Superpixel-Guided Correction incorporates a noise filtering mechanism that effectively suppresses erroneous information propagation in affinity modeling.
\subsection{Cross-Modal Prototype Alignment}
\subsubsection{Multimodal Prototype Generation}To address the inherent semantic gap between vision-language modalities, we propose a dual-branch alignment solution: For a batch of $N$ image-text pairs $\{(I_i, T_i)_{i=1}^{N}\}$, the structurally identical yet parameter-independent Image Semantic Alignment (ISA) and Text Semantic Alignment (TSA) modules refine the visual features $v_i' \in \mathbb{R}^{1 \times d_1}$ and textual features $t_i' \in \mathbb{R}^{1 \times d_1}$ extracted by CLIP, respectively. With customized loss function constraints, this solution significantly improves fine-grained semantic alignment of cross-modal features. Higher semantically aligned image and text representations are obtained as:
\begin{equation}
\label{deqn_ex1a}
v_i' = \text{ISA}(v_i), \quad t_i' = \text{TSA}(t_i) ,
\end{equation}
where $v_i' \in \mathbb{R}^{1 \times d_2}$ and $t_{i}' \in \mathbb{R}^{c_{f} \times d_{2}}$ denote the projected image features and text features obtained through the ISA module and TSA module, respectively, where $c_f$ represents the number of categories for the current image.

For each image, we generate $CAM_{ij}^c$ using GradCAM\cite{selvaraju2017grad} and compute image/text-specific prototypes. These prototypes are constructed in the projected space of ISA and TSA for two key reasons: (1) to preserve CLIP's inherent instance discrimination capability by shifting prototype discrimination to the projected space, and (2) to significantly reduce prototype construction costs through dimensionality reduction via the projection heads in these modules. Specifically, the image features $f_{image}$ and text features $f_{text}$ containing foreground target information are computed by:
\begin{equation}
\label{deqn_ex1a}
f_{image} = MAP(CAM^c \odot v_i'), \quad f_{text} = t_i'[index],
\end{equation}

where $MAP(\cdot)$ represents masked average pooling. Since the text feature $t_i'$ has dimensions ${R}^{c_{f} \times d_{2}}$, the text feature for the target category can be directly retrieved using the index.

We then collect the foreground-aware image features $f_{image}$ and text features $f_{text}$ from all image-text pairs in the dataset and perform K-means clustering to obtain the image prototypes $P^I \in \mathbb{R}^{K \times d_2} = [P_1^I, P_2^I, \dots, P_K^I]$ and text prototypes $P^T \in \mathbb{R}^{K \times d_2} = [P_1^T, P_2^T, \dots, P_K^T]$. After K-means clustering, cluster pseudo-labels can be generated based on the proximity between each sample's representation and its corresponding prototype.

\subsubsection{Prototype Contrastive Learning}This study proposes a prototype contrastive learning, which achieves fine-grained semantic alignment through triple constraints: 1) visual features are matched with text prototypes of the same category; 2) text prototypes are aggregated with visual prototypes of the same category; and 3) cross-modal prototypes of different categories are separated. As shown in Figure \ref{fig2} (a), this design guides image features toward semantic content while enabling text features to focus on visually alignable attributes, directly narrowing the modality gap and enhancing inter-class discriminability in a shared embedding space, thereby effectively addressing the issue of category confusion. The core innovation lies in constructing cross-modal positive and negative sample pairs, which synchronously optimizes modality alignment and classification boundaries through contrastive learning. To establish contrastive learning, the positive and negative sample pairs constructed using visual features $v_i'$ and text prototypes $P^T$ are defined as:
\begin{equation}
\label{deqn_ex1a}
p_i^I = \frac{v_i' \cdot P^T}{\tau_{proto}},
\end{equation}
\begin{equation}
\mathcal{S}_{pos}^i = \left\{ p_{i,j} \mid j = pos_{idx} \right\},
\end{equation}
\begin{equation}
\left\{ \mathcal{S}_{neg_1}^i, \dots, \mathcal{S}_{neg_k}^i \right\} = \left\{ p_{i,j} \mid j \neq pos_{idx} \right\},
\end{equation}
where the temperature hyperparameter $\tau_{proto}$ is set as a learnable parameter to optimize model performance. For constructing contrastive learning samples, $pos_{idx}$ denotes the cluster-generated pseudo-labels that guide the formation of positive and negative sample pairs, where $\mathcal{S}_{pos}^i$ represents the positive sample pair and $\mathcal{S}_{neg_k}^i$ indicates the$ k$-$th$ negative sample pair. Based on this, we employ the cross-entropy loss function to compute the average loss of all in-batch samples, and the prototype contrastive loss $\mathcal{L}_{proto}$ is implemented by:
\begin{equation}
\label{deqn_ex1a}
\mathcal{L}_{proto} = -\left( \frac{1}{N} \sum_{i=1}^{N} \log  \frac{\exp(S_{pos}^i)}{\exp(S_{pos}^i) + \sum_{j=1}^{k} \exp(S_{neg_j}^i)}  \right).
\end{equation}

\subsection{Superpixel-Guided Correction}
\subsubsection{Superpixel Clustering}
To address the issue of erroneous affinity propagation in attention mechanisms leading to the misactivation of background regions, we propose a Superpixel-Guided Correction. As shown in the Figure \ref{fig2}(b), this module leverages superpixel structural information to construct binary masks, selectively masking column vectors in the affinity matrix associated with non-target regions, thereby effectively suppressing the propagation of erroneous semantics in background regions. By constraining the attention propagation range with structured priors, only semantic correlations within target regions are preserved. Specifically, we define a binary mask matrix $Mask$, whose elements are defined as:
\begin{equation}
Mask = 
\begin{cases} 
a_{ij} = 1  & \text{if } j \in \textit{target regions}, \\ 
a_{ij} = 0  & \text{if } j \notin \textit{target regions},
\end{cases}
\end{equation}
where $i$ and $j$ represent the row and column indices of the affinity matrix respectively. 

To more accurately extract the target regions of the input image $I_i$, we employ the SLIC\cite{achanta2012slic} algorithm for superpixel segmentation. This method clusters pixels based on feature similarity, efficiently representing the image with a reduced number of superpixels while preserving object boundary integrity. Subsequently, the superpixel regions are clustered based on color space information to obtain the target region $C$ as:
\begin{equation}
C = \text{K-means}(\text{SLIC}(I_i)).
\end{equation}

We calculate the ratio of the sum of high-confidence pixel activations to the total activation value within each clustering region. Only regions with a ratio above a predefined threshold are classified as \textit{target regions}. Additionally, the lightweight design of SLIC makes it more suitable than complex models like SAM for the role of spatial prior in SGC.

\subsubsection{Affinity Matrix Correction}
CLIP's MHSA excels at extracting global semantic features, but its insufficient spatial detail capture leads to blurred boundaries in CAM. In contrast, DINO's MHSA, enhanced through self-supervised training, strengthens local-to-global consistency. We use DINO's local structures to improve CAM spatial priors. To address this, we integrate their MHSA features and normalize them to obtain the affinity matrix $A$: CLIP provides high-level semantic guidance, while DINO supplements fine-grained spatial relationships, thereby generating CAM that maintain category discriminability while achieving precise spatial localization, with the fused affinity matrix computed as:
\begin{equation}
A = ConCat(MHSA_{CLIP}, MHSA_{DINO}),
\end{equation}
where $MHSA_{CLIP}$ represents the MHSA from CLIP and $MHSA_{DINO}$ denotes the MHSA from DINO, the affinity matrix $A$ is obtained by normalizing their $ConCat$ operation, effectively combining both attention mechanisms for cross-modal feature integration.

Subsequently, we refine the affinity matrix $A$ using the obtained $Mask$ to derive an updated matrix $A^*$, where redundant non-target column elements are eliminated. This refined affinity matrix $A^*$ is then employed to enhance the initial CAM through spatial propagation. The rectified affinity matrix $A^*$ and the final CAM are computed as:
\begin{equation}
A^* = A \odot Mask,
\end{equation}
\begin{equation}
CAM_{refine}^c = A^* \otimes CAM^c,
\end{equation}
where $CAM_{refine}^c$ denotes the refined CAM corresponding to foreground target class $c$.

\begin{table}[]
\centering
\begin{tabular}{cccc}
\toprule
\multicolumn{1}{l|}{\multirow{2}{*}{Method}} & \multicolumn{2}{c|}{VOC}                              & COCO                 \\ \cline{2-4} 
\multicolumn{1}{c|}{}                        & Val                      & \multicolumn{1}{c|}{Test}  & Val                  \\ \midrule
\multicolumn{4}{l}{\textit{\textbf{Multi-stage.}}}                                                                                                                                                       \\
\multicolumn{1}{l|}{SIPE~\cite{chen2022self}}          & 68.8                     & \multicolumn{1}{c|}{69.7}  & 43.6                 \\
\multicolumn{1}{l|}{CLIMS~\cite{xie2022clims}}         & 70.4                     & \multicolumn{1}{c|}{70.0}  & -                    \\
\multicolumn{1}{l|}{WeakTr~\cite{zhu2023weaktr}}      & 78.4                     & \multicolumn{1}{c|}{79.0}  & 50.3                 \\
\multicolumn{1}{l|}{CLIP-ES~\cite{lin2023clip}}      & 73.8                     & \multicolumn{1}{c|}{73.9}  & 45.4                 \\
\multicolumn{1}{l|}{PSDPM~\cite{zhao2024psdpm}}         & 74.1                     & \multicolumn{1}{c|}{74.9}  & 47.2                 \\
\multicolumn{1}{l|}{CPAL~\cite{tang2024hunting}}          & 74.5                     & \multicolumn{1}{c|}{74.7}  & 46.8                 \\
\multicolumn{1}{l|}{CTI~\cite{yoon2024class}}           & \multicolumn{1}{l}{74.1} & \multicolumn{1}{l|}{73.2}  & 45.4                 \\
\multicolumn{1}{l|}{WeakCLIP~\cite{zhu2025weakclip}}      & 74.0                     & \multicolumn{1}{c|}{73.8}  & 47.4                 \\
\multicolumn{1}{l|}{VPL~\cite{xu2025toward}}           & 79.3                     & \multicolumn{1}{c|}{79.0}  & 49.8                 \\ \bottomrule
\multicolumn{4}{l}{\textit{\textbf{Single-stage.}}}                                                                                                                             \\
\multicolumn{1}{l|}{DIAL~\cite{jang2024dial}}          & 74.5                     & \multicolumn{1}{c|}{74.9}  & 44.4                 \\
\multicolumn{1}{l|}{DuPL~\cite{wu2024dupl}}          & 73.3                     & \multicolumn{1}{c|}{72.8}  & 44.6                 \\
\multicolumn{1}{l|}{WeCLIP~\cite{zhang2024frozen}}       & 76.4                     & \multicolumn{1}{c|}{77.2}  & 47.1                 \\
\multicolumn{1}{l|}{MoRe~\cite{yang2025more}}          & 76.4                     & \multicolumn{1}{c|}{75.0}  & 47.4                 \\
\multicolumn{1}{l|}{ExCEL~\cite{yang2025exploring}}        & 78.4                     & \multicolumn{1}{c|}{78.5}  & 50.3                 \\
\rowcolor{gray!15}
\multicolumn{1}{l|}{\textbf{Ours w/o CRF}}            & \textbf{78.2} & \multicolumn{1}{c|}{\textbf{78.1}} & \textbf{49.2}                \\
\rowcolor{gray!15}
\multicolumn{1}{l|}{\textbf{Ours}}                    & \textbf{79.5}                    & \multicolumn{1}{c|}{\textbf{79.6}} & \textbf{50.6}                 \\ \bottomrule
\end{tabular}
\caption{\protect\normalfont Segmentation performance comparison (mIoU\%) on PASCAL VOC and COCO datasets, showing results on VOC val/test sets and COCO val set.}
\label{table1}
\end{table}

\begin{table}[]
\centering
\begin{tabular}{cccc}
\toprule
\multicolumn{1}{l|}{\multirow{2}{*}{Method}} & \multicolumn{1}{c|}{\multirow{2}{*}{Sup.}} & \multicolumn{1}{c|}{\multirow{2}{*}{N.}} & \multicolumn{1}{c}{VOC} \\ \cline{4-4} 
\multicolumn{1}{c|}{} & \multicolumn{1}{c|}{} & \multicolumn{1}{c|}{} & Train \\ \midrule
\multicolumn{1}{l|}{MCTformer {\scriptsize CVPR'2022}} & \multicolumn{1}{c|}{$\mathcal{I + D}$} & \multicolumn{1}{c|}{R} & 61.7 \\
\multicolumn{1}{l|}{CLIMS~\cite{xie2022clims}} & \multicolumn{1}{c|}{$\mathcal{I + L}$} & \multicolumn{1}{c|}{R} & 56.6 \\
\multicolumn{1}{l|}{WeakTr~\cite{zhu2023weaktr}} & \multicolumn{1}{c|}{$\mathcal{I}$} & \multicolumn{1}{c|}{D} & 66.2 \\
\multicolumn{1}{l|}{POLE {\scriptsize WACV'2023}} & \multicolumn{1}{c|}{$\mathcal{I + L}$} & \multicolumn{1}{c|}{R} & 59.0 \\
\multicolumn{1}{l|}{CLIP-ES~\cite{lin2023clip}} & \multicolumn{1}{c|}{$\mathcal{I + L}$} & \multicolumn{1}{c|}{V} & 70.8 \\
\multicolumn{1}{l|}{ToCo~\cite{ru2023token}} & \multicolumn{1}{c|}{$\mathcal{I}$} & \multicolumn{1}{c|}{V} & 71.6 \\
\multicolumn{1}{l|}{CPAL~\cite{tang2024hunting}} & \multicolumn{1}{c|}{$\mathcal{I + L}$} & \multicolumn{1}{c|}{R} & 71.9 \\
\multicolumn{1}{l|}{CTI~\cite{yoon2024class}} & \multicolumn{1}{c|}{$\mathcal{I}$} & \multicolumn{1}{c|}{D} & 69.5 \\
\multicolumn{1}{l|}{SeCo~\cite{yang2024separate}} & \multicolumn{1}{c|}{$\mathcal{I}$} & \multicolumn{1}{c|}{V} & 74.8 \\ 
\multicolumn{1}{l|}{DuPL~\cite{wu2024dupl}} & \multicolumn{1}{c|}{$\mathcal{I}$} & \multicolumn{1}{c|}{V} & 75.0 \\
\multicolumn{1}{l|}{WeCLIP~\cite{zhang2024frozen}} & \multicolumn{1}{c|}{$\mathcal{I + L}$} & \multicolumn{1}{c|}{V} & 75.4 \\
\multicolumn{1}{l|}{DIAL~\cite{jang2024dial}} & \multicolumn{1}{c|}{$\mathcal{I + L}$} & \multicolumn{1}{c|}{V} & 75.2 \\
\multicolumn{1}{l|}{WeakCLIP~\cite{zhu2025weakclip}} & \multicolumn{1}{c|}{$\mathcal{I + L}$} & \multicolumn{1}{c|}{V} & 61.7 \\
\multicolumn{1}{l|}{VPL~\cite{xu2025toward}} & \multicolumn{1}{c|}{$\mathcal{I + L}$} & \multicolumn{1}{c|}{V} & 77.8 \\
\multicolumn{1}{l|}{MoRe~\cite{yang2025more}} & \multicolumn{1}{c|}{$\mathcal{I}$} & \multicolumn{1}{c|}{V} & 77.0 \\
\multicolumn{1}{l|}{ExCEL~\cite{yang2025exploring}} & \multicolumn{1}{c|}{$\mathcal{I + L}$} & \multicolumn{1}{c|}{V} & 78.0 \\
\rowcolor{gray!15}
\multicolumn{1}{l|}{\textbf{Ours}} & \multicolumn{1}{c|}{$\mathcal{I+L+D}$} & \multicolumn{1}{c|}{V} & \textbf{78.7} \\ \bottomrule
\end{tabular}
\caption{\protect\normalfont CAM seed comparisons on VOC train set. D: DINO. N: network. R: ResNet. D:Deit. V:Vit-B.}
\label{table2}
\end{table}

\subsection{Training Objectives}
The overall loss of SSR consists of two components: a prototype contrastive loss $\mathcal{L}_{proto}$ and a segmentation loss $\mathcal{L}_{seg}$. $\mathcal{L}_{proto}$ encourages samples to move closer to their corresponding cross-modal prototypes of the same class while pushing them away from prototypes of different classes. $\mathcal{L}_{seg}$ employs online-generated pseudo masks and adopts a cross-entropy formulation for end-to-end training to accomplish the segmentation task. The objective function of our approach can be expressed as:
\begin{equation}
\mathcal{L}_{SSR} = \mathcal{L}_{proto} + \gamma \mathcal{L}_{seg}.
\end{equation}

\section{Experiments}
\subsection{Experimental Settings}
\subsubsection{Datasets and Metrics}The proposed method SSR is evaluated on both PASCAL VOC 2012\cite{everingham2015pascal} and MS COCO 2014\cite{lin2014microsoft} datasets, where VOC comprises 21 categories (including 1 background class) and follows the established protocol\cite{li2021group, du2022weakly} using augmented sets of 10,582 training images, 1,449 validation images, and 1,456 test images. COCO contains 81 categories with 82,081 training images and 40,137 validation images. We employ mIoU as the primary evaluation metric, complemented by secondary metrics including confusion ratio, precision (P) and recall (R) for comprehensive performance assessment.

\subsubsection{Implementation Details}CLIP adopts the ViT-B/16, while the DINO utilizes the ViT-S/16 architecture from DINOv1. We used the AdamW optimizer with a learning rate of 1e-5 and weight decay of 2e-3. For PASCAL VOC 2012, we set the batch size to 128 with a maximum of 30,000 iterations, whereas MS COCO-2014 uses a larger batch size of 256 and 80,000 iterations. The loss weight is set to 0.1 with a prototype temperature coefficient $\tau_{\text{Proto}}$=0.05. Both ISA and TSA consist of stacked linear layers with batch normalization and ReLU. The prototypes are updated every 5,000 iterations. In SGC, the CLIP:DINO weighting ratio is 0.4:0.6.

\subsection{Comparisons with State-of-the-art Methods}
\subsubsection{Performance of Semantic Segmentation}
\begin{figure*}[t] 
\centering
\includegraphics[width=\textwidth]{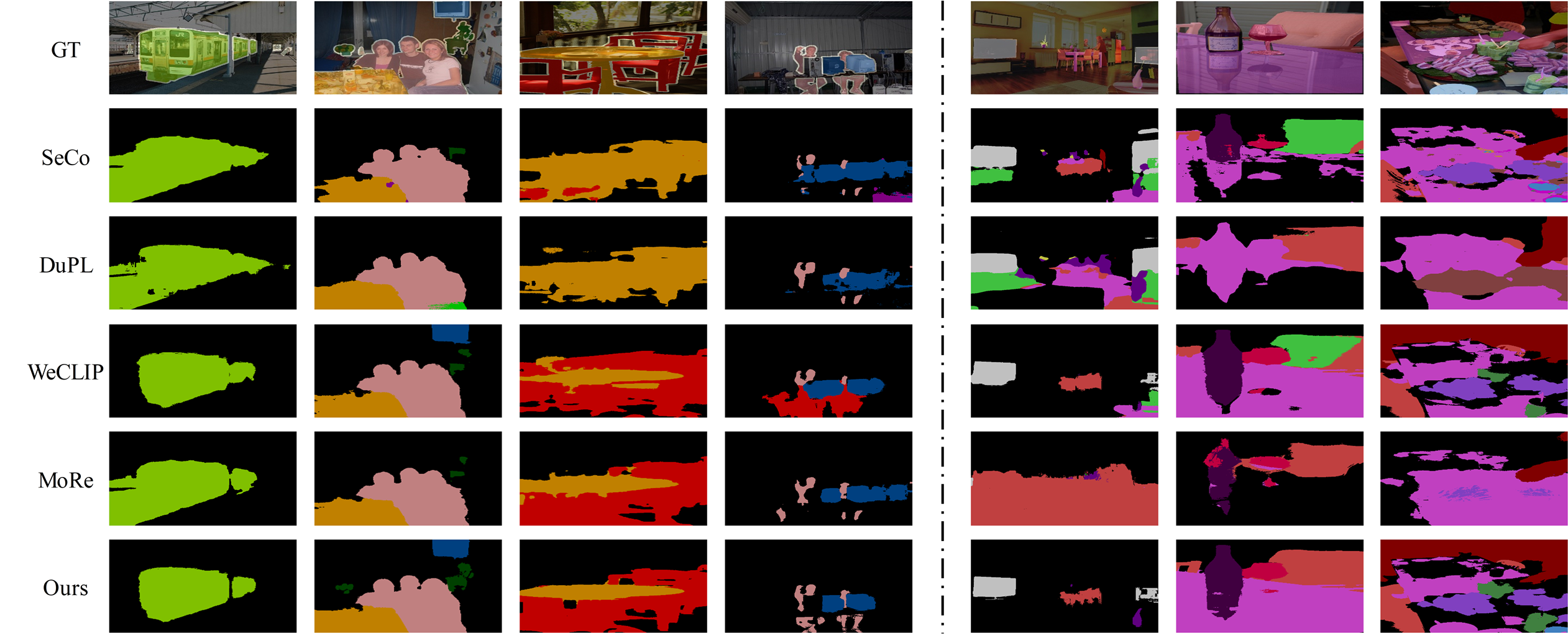} 
\caption{Segmentation visualizations of SeCo, DUPL, WeCLIP, MoRe, and Ours on VOC and COCO. Columns 1-4: Results on PASCAL VOC dataset. Columns 5-7: Results on the MS COCO dataset. SSR segments objects more precisely.}
\label{fig5} 
\end{figure*}

Table \ref{table1} presents a comparison of segmentation performance between SSR and recent methods on the VOC and COCO datasets. SSR achieves new SOTA performance on both VOC and COCO, surpassing multi-stage methods by up to 0.6\% and CLIP-based ExCEL by 0.3\%. As shown in Figure \ref{fig5}, qualitative results demonstrate superior segmentation quality through: (1) enhanced cross-modal contrastive learning for accurate category prediction; (2) improved region integrity and boundary clarity; and (3) stronger inter-class discriminability in multi-category scenes.

\begin{figure}[t]
\centering
\includegraphics[width=0.98\columnwidth]{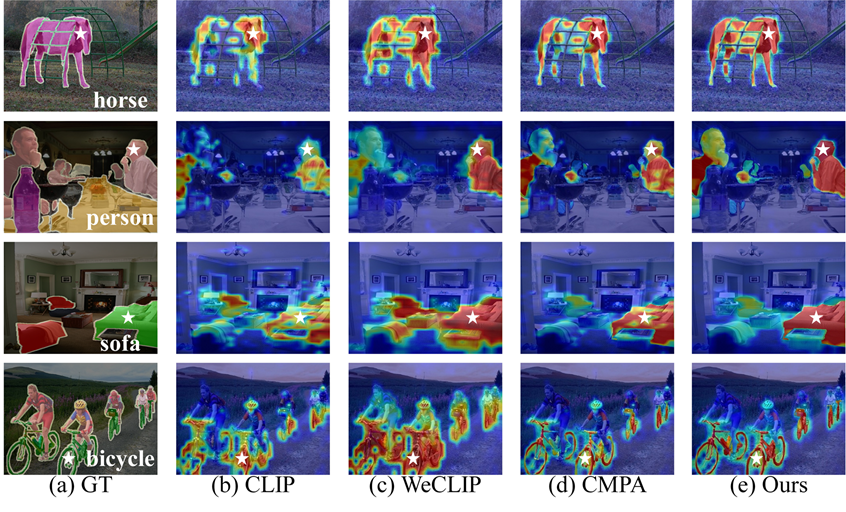}
\caption{CAM visualizations on VOC val set.We conduct a comparative analysis between the initial CAMs generated by CLIP and those produced by our CMPA, followed by evaluating the performance gap between WeCLIP's results and our final optimized outputs.}
\label{fig6}
\end{figure}

\begin{table}[t]
\begin{tabularx}{\columnwidth}{>{\centering\arraybackslash}X|cccccc}
\toprule
CPCM & $At_{\text{clip}}$ & $At_{\text{dino}}$ & SGC & P    & R    & mIoU \\ \midrule
\checkmark    &                 &                  &     & 72.8 & 84.6 & 63.3 \\
\checkmark   & \checkmark               &                  &     & 85.2 & 88.9 & 74.6 \\
\checkmark    & \checkmark                 & \checkmark                &     & 84.3 & 86.2 & 76.3 \\
\rowcolor{gray!15}
\checkmark    & \checkmark                 & \checkmark                  & \checkmark   & \textbf{87.9} & \textbf{89.1} & \textbf{78.7} \\ 
\bottomrule
\end{tabularx}
\caption{\protect\normalfont Ablation study of SGC on the PASCAL VOC train set. At denotes attention map.}
\label{table3}
\end{table}

\begin{table}[h]
\centering
\begin{tabularx}{\columnwidth}{>{\centering\arraybackslash}X|ccccc}
\toprule
Conditions        & $L_{\text{feature}}$ & $L_{\text{in\_modal}}$ & $L_{\text{cross\_modal}}$ & mIoU \\ 
\midrule
CLIP    &                      &                        &                           & 58.6 \\
w $L_{\text{feature}}$      & \checkmark         &                        &                           & 53.5 \\
w $L_{\text{in\_modal}}$    &                    & \checkmark             &                           & 57.8 \\
\rowcolor{gray!15}
w $L_{\text{cross\_modal}}$ &                    &                        & \checkmark                & \textbf{63.3} \\  
\bottomrule
\end{tabularx}
\caption{\protect\normalfont Performance evaluation of SSR using three different loss functions}
\label{table4}
\end{table}

\subsubsection{Evaluation of CAM Seeds}Table \ref{table2} reports the quality of CAM seeds on the VOC training set. Compared to recent methods, SSR further improves the CAM quality to 78.7\%, surpassing the SOTA by at least 0.7\%. As shown in Figure \ref{fig6}, the CAM visualization results demonstrate that the initial CAMs generated by CMPA show significant advantages over CLIP, exhibiting superior completeness. Taking the "sofa" category as an example, the CLIP method erroneously activates adjacent "chair" regions, and this error gets amplified during subsequent processing. SSR not only achieves more precise focus on target regions but also show stronger background noise suppression capability.

\subsection{Ablation Study}
\subsubsection{Effectiveness of the SGC}Table \ref{table3} presents the component-wise ablation study of SGC on PASCAL VOC: The CPMA-generated initial CAM achieves 63.3\% mIoU, which improves to 74.6\% after incorporating CLIP's multi-head attention. Further integration with DINO's attention mechanism yields 76.3\% mIoU, demonstrating its complementary role in enhancing inter-patch semantic relations. Our complete SGC module ultimately achieves 78.7\% mIoU by effectively rectifying fusion errors in attention maps during CAM refinement.

\begin{table}[]
\centering
\begin{tabularx}{\columnwidth}{>{\centering\arraybackslash}X|cccc}
\toprule
Method        & mIoU           & Precision      & Recall         & Confusion           \\ 
\midrule
SeCo          & 0.740          & 0.84           & 0.849          & 0.232          \\
WeCLIP        & 0.764          & 0.844          & 0.861          & 0.237          \\
MoRe          & 0.764          & 0.837          & 0.847          & 0.239          \\
\rowcolor{gray!15}
\textbf{Ours} & \textbf{0.795} & \textbf{0.879} & \textbf{0.891} & \textbf{0.198} \\
\bottomrule
\end{tabularx}
\caption{\protect\normalfont Comparison with recent methods on the PASCAL VOC val set in terms of four metrics.}
\label{table5}
\end{table}

\subsubsection{Effectiveness of the training loss}We evaluated three different loss functions on the PASCAL VOC dataset. As shown in Table \ref{table4}, baseline CLIP achieves 58.6\% mIoU, while direct fine-tuning($L_{\text{feature}}$) degrades performance by 5.1\%; Intra-modal contrastive learning($L_{\text{in\_modal}}$) causes only a marginal 0.8\% drop, indicating limited efficacy of single-modal prototype comparison; In contrast, our cross-modal contrastive loss($L_{\text{cross\_modal}}$) significantly improves mIoU by 4.7\% by aligning cross-modal representations of the same class while separating different classes, mitigating the modal semantic gap.

\subsubsection{Model Performance Across Key Metrics}Table \ref{table5} compares our method with SeCo, MoRe and WeCLIP on PASCAL VOC val set across four key metrics. Our method achieves superior performance on all indicators: surpassing the best baseline MoRe by 3.1\% in mIoU and 3.5\% in Precision, demonstrating enhanced detection accuracy. The 3\% Recall improvement indicates more complete segmentation with fewer false activations, while the 3.4\% reduction in Confusion ratio confirms better inter-class discrimination.

\begin{table}[]
\centering
\begin{tabular}{l|c|c|c|c}
\toprule
\multicolumn{1}{l|}{\multirow{2}{*}{Methods}} & \multicolumn{1}{c|}{\multirow{2}{*}{Sup.}} & \multicolumn{1}{c|}{\multirow{2}{*}{Net.}} & \multicolumn{1}{c|}{\multirow{2}{*}{Val}} & \multicolumn{1}{c}{\multirow{2}{*}{Ratio}} \\ 
\multicolumn{1}{l|}{} & \multicolumn{1}{c|}{} & \multicolumn{1}{c|}{} & \multicolumn{1}{c|}{} & \multicolumn{1}{c}{} \\ \midrule
\multicolumn{1}{l|}{DeepLabV2} & \multicolumn{1}{c|}{F} & \multicolumn{1}{c|}{RN101} & \multicolumn{1}{c|}{77.7} & \multicolumn{1}{c}{-} \\
\multicolumn{1}{l|}{DeepLabV2} & \multicolumn{1}{c|}{F} & \multicolumn{1}{c|}{ViT-B} & \multicolumn{1}{c|}{82.3} & \multicolumn{1}{c}{-} \\
\midrule
\multicolumn{1}{l|}{CLIMS {\scriptsize CVPR'2022}} & \multicolumn{1}{c|}{$\mathcal{I + L}$} & \multicolumn{1}{c|}{RN101} & \multicolumn{1}{c|}{70.4} & \multicolumn{1}{c}{90.6\%} \\
\multicolumn{1}{l|}{CLIP-ES {\scriptsize CVPR'2023}} & \multicolumn{1}{c|}{$\mathcal{I + L}$} & \multicolumn{1}{c|}{RN101} & \multicolumn{1}{c|}{72.2} & \multicolumn{1}{c}{92.9\%} \\
\multicolumn{1}{l|}{CPAL {\scriptsize CVPR'2024}} & \multicolumn{1}{c|}{$\mathcal{I + L}$} & \multicolumn{1}{c|}{RN101} & \multicolumn{1}{c|}{74.5} & \multicolumn{1}{c}{95.9\%} \\
\multicolumn{1}{l|}{ToCo {\scriptsize CVPR'2024}} & \multicolumn{1}{c|}{$\mathcal{I}$} & \multicolumn{1}{c|}{ViT-B} & \multicolumn{1}{c|}{71.1} & \multicolumn{1}{c}{86.4\%} \\
\multicolumn{1}{l|}{DuPL {\scriptsize CVPR'2024}} & \multicolumn{1}{c|}{$\mathcal{I}$} & \multicolumn{1}{c|}{ViT-B} & \multicolumn{1}{c|}{73.3} & \multicolumn{1}{c}{89.1\%} \\
\multicolumn{1}{l|}{SeCo {\scriptsize CVPR'2024}} & \multicolumn{1}{c|}{$\mathcal{I}$} & \multicolumn{1}{c|}{ViT-B} & \multicolumn{1}{c|}{74.0} & \multicolumn{1}{c}{89.9\%} \\
\multicolumn{1}{l|}{DIAL {\scriptsize ECCV'2024}} & \multicolumn{1}{c|}{$\mathcal{I + L}$} & \multicolumn{1}{c|}{ViT-B} & \multicolumn{1}{c|}{74.5} & \multicolumn{1}{c}{90.5\%} \\
\multicolumn{1}{l|}{WeCLIP {\scriptsize CVPR'2024}} & \multicolumn{1}{c|}{$\mathcal{I + L}$} & \multicolumn{1}{c|}{ViT-B*} & \multicolumn{1}{c|}{76.4} & \multicolumn{1}{c}{93.6\%} \\
\multicolumn{1}{l|}{ExCEL {\scriptsize CVPR'2025}} & \multicolumn{1}{c|}{$\mathcal{I + L}$} & \multicolumn{1}{c|}{ViT-B*} & \multicolumn{1}{c|}{78.4} & \multicolumn{1}{c}{96.1\%} \\
\rowcolor{gray!15}
\multicolumn{1}{l|}{\textbf{Ours}} & \multicolumn{1}{c|}{$\mathcal{I + L}$} & \multicolumn{1}{c|}{\textbf{ViT-B*}} & \multicolumn{1}{c|}{\textbf{79.5}} & \multicolumn{1}{c}{\textbf{97.4\%}} \\
\bottomrule
\end{tabular}
\caption{\protect\normalfont Comparisons with the fully-supervised methods on VOC val set. F:fully-supervised. ViT-B*: pretrained from CLIP.}
\label{table6}
\end{table}

\subsubsection{Fully-supervised Counterparts}Table \ref{table6} systematically compares SSR with fully-supervised approaches. The smaller performance gap demonstrates our method's effectiveness. SSR achieves 79.5\% mIoU on VOC 2012 val set, reaching 97.4\% of fully-supervised performance. The results demonstrate significant advantages over existing WSSS methods.

\begin{figure}[!t]
\centering
\includegraphics[width=0.98\columnwidth]{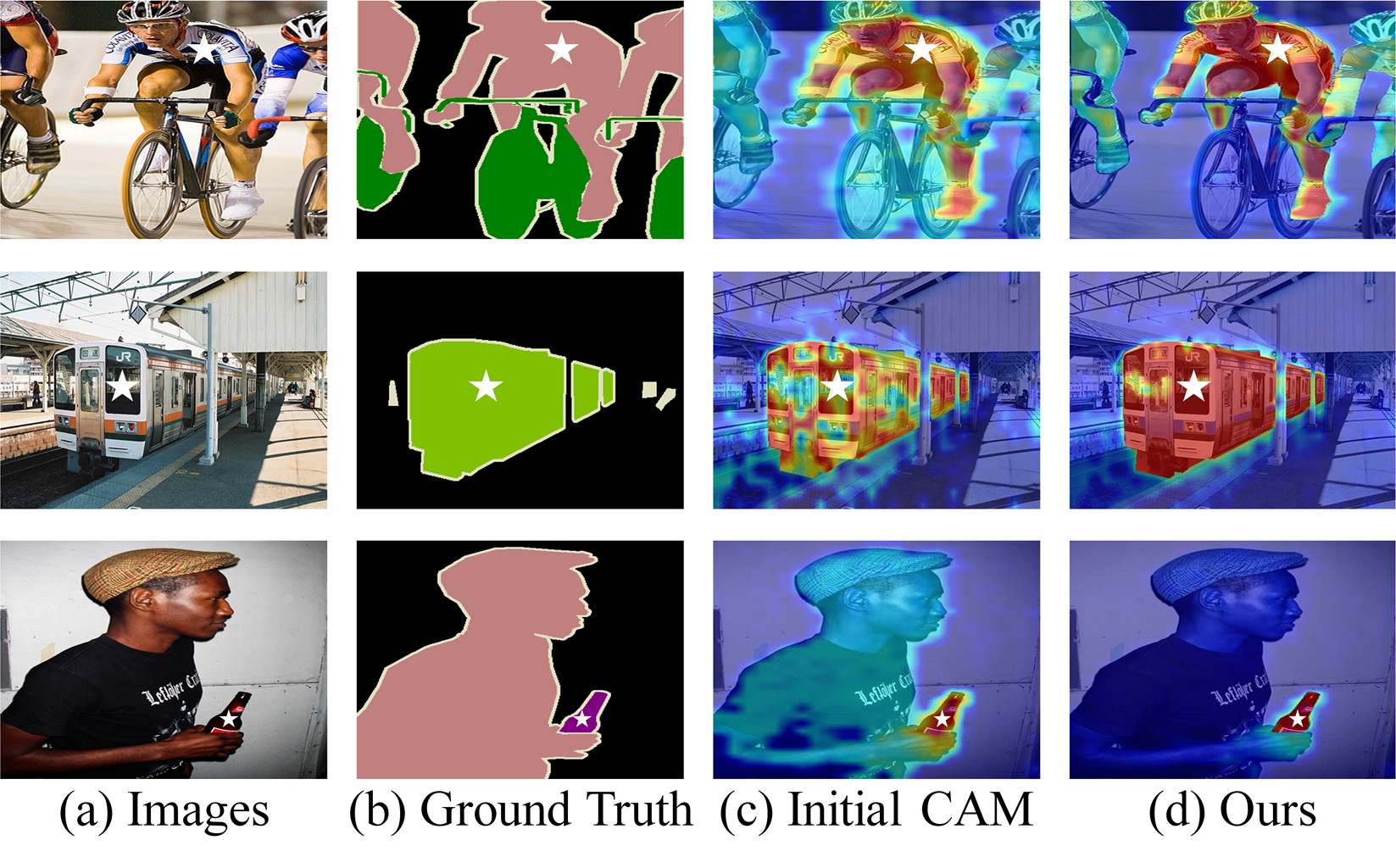}
\caption{Visualization of CAM refinement: (c) Initial CAM with background artifacts; (d) Refined CAM after SGC processing, showing cleaner background suppression and sharper target focus.}
\label{fig7}
\end{figure}

\subsubsection{Visualization of CAM refinement}Figure \ref{fig7} illustrates the optimization effect of the SGC on the initial CAM. After superpixel refinement, the erroneous activations in background regions are significantly suppressed. This improvement primarily benefits from the carefully designed affinity matrix, which effectively eliminates interference from affinity relationships in non-target regions, thereby achieving more precise localization of activated areas.

\begin{figure}[t]
\centering
\includegraphics[width=0.98\columnwidth]{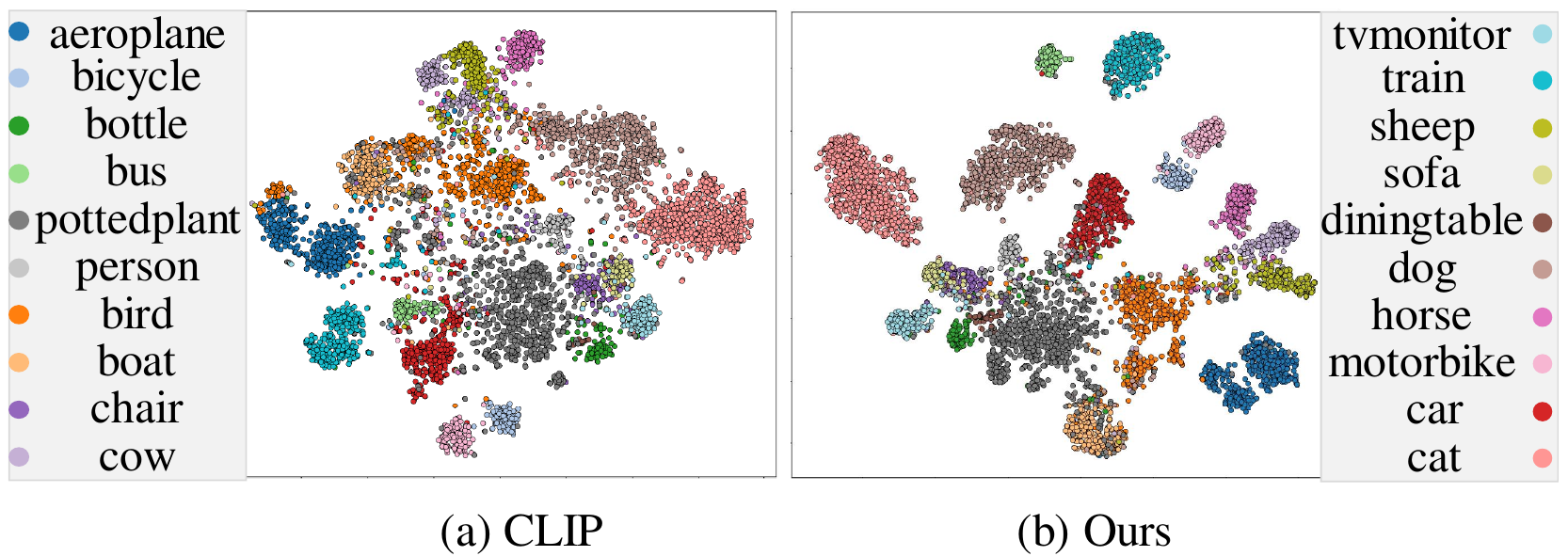}
\caption{The t-SNE\cite{van2008visualizing} visualization of feature embeddings on Pascal VOC 2012 validation images shows: (a) the original CLIP feature space distribution, and (b) the optimized feature distribution.}
\label{fig8}
\end{figure}

\subsubsection{Effectiveness of feature alignment}To validate CPMA's effectiveness, Figure \ref{fig8} compares CLIP features with CPMA-enhanced features on PASCAL VOC 2012. While CLIP features show good transferability but sparse distribution with class overlap, CPMA produces more compact intra-class clusters and clearer inter-class boundaries (e.g., distinct diningtable/bottle separation).

\section{Conclusion}
We proposes SSR to address the issues of modality gap and spurious background responses in CLIP-based WSSS, effectively suppressing erroneous activations in non-target foreground and background regions. Specifically, our proposed CMPA establishes a contrastive relationship between features and cross-modal prototypes, achieving intra-class aggregation and inter-class separation. Meanwhile, SGC dynamically adjusts the propagation direction of feature affinity to effectively suppress redundant correlations in non-target regions within the affinity matrix. Extensive experiments demonstrate that our method exhibits significant advantages in segmentation accuracy and error suppression, fully validating the effectiveness of the proposed solution.

\section{Acknowledgements}
This work was supported in part by the National Science Foundation of China Joint Key Project under Grants U24B20173, and by the National Natural Science Foundation of China under Grant 62536002, Grant 62561160098, Grant 62576055 and Grant 32350009.

\bibliography{aaai2026}

\end{document}